\documentclass[10pt]{article} 
\usepackage[accepted]{tmlr}

\usepackage{amsmath,amsfonts,bm}

\def\eqref#1{equation~\ref{#1}}

\def\1{\bm{1}}

\DeclareMathAlphabet{\mathsfit}{\encodingdefault}{\sfdefault}{m}{sl}
\SetMathAlphabet{\mathsfit}{bold}{\encodingdefault}{\sfdefault}{bx}{n}

\usepackage{hyperref}
\usepackage{url}
\usepackage{xspace}
\usepackage{graphicx}
\usepackage{enumitem}
\usepackage{multirow}
\usepackage{booktabs}
\usepackage{subfigure}
\usepackage{algorithm}
\usepackage{algorithmic}
\usepackage{bbding}
\usepackage{bm}

\title{Enhancing Fairness in Unsupervised Graph Anomaly Detection through Disentanglement}

\author{\name Wenjing Chang \email changwenjing@cnic.cn \\
      \addr Computer Network Information Center, Chinese Academy of Sciences \\ University of Chinese Academy of Sciences
      \AND
      \name Kay Liu \email zliu234@uic.edu \\
      \addr University of Illinois Chicago
      \AND
      \name Philip S. Yu \email psyu@uic.edu\\
      \addr University of Illinois Chicago 
      \AND
      \name Jianjun Yu \email yujj@cnic.ac.cn\\
      \addr Computer Network Information Center, Chinese Academy of Sciences }

\newcommand{\method}{{DEFEND}\xspace}

\def\month{03}  
\def\year{2025} 
\def\openreview{\url{https://openreview.net/forum?id=5zRs34Ls3C}} 

\begin{document}

\maketitle

\begin{abstract}
Graph anomaly detection (GAD) is becoming increasingly crucial in various applications, ranging from financial fraud detection to fake news detection.
However, current GAD methods largely overlook the fairness problem, which might result in discriminatory decisions skewed toward certain demographic groups defined on sensitive attributes (e.g., gender). 
This greatly limits the applicability of these methods in real-world scenarios in light of societal and ethical restrictions. 
To address this critical gap, we make the first attempt to integrate fairness with utility in GAD decision-making.
Specifically, we devise a novel \textbf{D}is\textbf{E}ntangle-based \textbf{F}airn\textbf{E}ss-aware a\textbf{N}omaly \textbf{D}etection framework on the attributed graph, named \textbf{\method}.
\method first introduces disentanglement in GNNs to capture informative yet sensitive-irrelevant node representations, effectively reducing bias inherent in graph representation learning. 
Besides, to alleviate discriminatory bias in evaluating anomalies, \method adopts a reconstruction-based method, which concentrates solely on node attributes and avoids incorporating biased graph topology.
Additionally, given the inherent association between sensitive-relevant and -irrelevant attributes, \method further constrains the correlation between the reconstruction error and predicted sensitive attributes.
Empirical evaluations on real-world datasets reveal that \method performs effectively in GAD and significantly enhances fairness compared to state-of-the-art baselines.
Our code is available at \url{https://github.com/AhaChang/DEFEND}.
\end{abstract}

\section{Introduction}
Graph Anomaly Detection (GAD), which aims to identify nodes that deviate significantly from the majority of nodes, has attracted wide attention in various domains, including fraudster detection in financial networks \citep{zhang2022efraudcom,huang2022dgraph} and spammer detection in social networks \citep{li2019spam,wu2020graph}.
The advancement of Graph Neural Networks (GNNs) \citep{kipf2016semi,hamilton2017inductive,velivckovic2017graph} has significantly 
enhanced the ability of GNN-based GAD methods to accurately identify anomalies \citep{ding2019deep,chai2022can,kim2023class,he2023ada}.
However, a recent study \citep{neo2024towards} reveal a concerning trend: current GAD methods exhibit substantial bias in decision.

Given the wide-ranging applications of GAD, particularly within high-stakes domains, the fairness problem cannot be overlooked.
Unfair decisions that skew toward certain demographic groups associated with sensitive attributes (e.g., gender, religion, ethnicity, etc.) might cause profound societal and ethical concerns. 
For example, in the realm of social networks (e.g., Reddit and Twitter), anomalous users (e.g., spreading misinformation or engaging in fake account interactions) might undergo strict investigation and even permanent account suspension.
In such scenarios, biased decisions could result in unfairly focusing on certain groups while inadvertently neglecting others. This undermines the effectiveness and reliability of anomaly detection systems and raises critical ethical concerns. 
To balance fairness and utility in anomaly detection, several methods have been proposed \citep{deepak2020fair,song2021deep,zhang2021towards,shekhar2021fairod}.
These methods strive to optimize the balance between fairness and anomaly detection performance in the absence of ground truth labels, which presents a fundamental challenge in unsupervised anomaly detection.
Nevertheless, they primarily focus on independent and identically distributed data, thereby overlooking the societal bias in graphs, which manifests in both node attributes and graph topology.

The bias in graphs poses a significant challenge to achieving fairness in graph-related tasks~\citep{dai2021say,zhu2023devil}.
First, sensitive attributes are inherently spread across other attributes \citep{deepak2020fair,oh2022learning}, so directly removing them is insufficient to ensure fairness \citep{neo2024towards}. For example, the geographic location might correlate with religion and ethnicity.
Second, since nodes with similar attributes are more likely to form connections, graph topology is also influenced by sensitive attributes~\citep{rahman2019fairwalk,spinelli2021fairdrop}.
Third, biased topology coupled with the message-passing mechanism in GNNs may inherit or even amplify the inherent bias in graphs~\citep{dai2021say,wang2022improving,zhu2023devil}.
Specifically, representations aggregated from neighboring nodes that share identical sensitive attributes may amplify features of the demographic group, thereby potentially affecting the fairness of the decision-making in GNNs.

Many efforts have been made to explore fairness for GNN-based methods~\citep{li2021dyadic,ma2022learning,kim2022debiasing,song2022guide}.
A common strategy involves eliminating sensitive information from the training graph and implementing the debiased graph for target tasks~\citep{spinelli2021fairdrop,rahman2019fairwalk,dong2022edits}. 
However, it is non-trivial to concurrently mitigate bias and preserve the integrity of anomalies, considering the overlap between the features of anomalies unrelated to sensitive attributes and the features of demographic groups linked to sensitive attributes.
For instance, an edge might indicate that two nodes share the same sensitive attributes while displaying an anomalous connection.
Another prevalent strategy involves training fair GNNs to perform the target task independently of sensitive attributes~\citep{dai2021say,zhu2023devil}.
These methods are dedicated to end-to-end supervised node classification, where ground-truth labels are available. 
Thus, applying them to unsupervised GAD is challenging due to the absence of labels for anomaly detection.

In this paper, we propose a novel \textbf{D}is\textbf{E}ntangle-based \textbf{F}airn\textbf{E}ss-aware a\textbf{N}omaly \textbf{D}etection framework on attributed graphs, named \textbf{\method}. 
In the first stage, to address the societal bias embedded in both node attributes and graph topology, we introduce disentangled fair representation learning on graphs to capture node representations that are both informative and independent of sensitive attributes. 
Specifically, the disentangled graph encoder can effectively separate sensitive-relevant and sensitive-irrelevant representations into independent subspaces with the guidance of a learnable adversary.
In the second stage, given the absence of ground truth labels and the inherent complex bias in graph topology, we implement an additional decoder that reconstructs node attributes from the well-trained disentangled encoder, utilizing the reconstruction error as the anomaly score.
To further alleviate discriminatory bias in detecting anomalies, we introduce a fairness constraint that enforces invariance of reconstruction errors across different sensitive attribute groups, effectively mitigating the influence of correlations between sensitive and the rest attributes.
Our main contributions are summarized as follows:
\begin{itemize}
    \item To the best of our knowledge, we proposed the first method \method for fair unsupervised graph anomaly detection, which reduces discriminatory bias in anomaly detection.
    \item \method employs constrained reconstruction error coupled with a disentangled graph encoder for fair anomaly detection on graphs.
    \item Extensive experiments on real-world datasets show that \method achieves a competitive performance and significantly enhances fairness compared with baselines.
\end{itemize}

\section{Preliminaries}
\subsection{Problem Definition}
Let $\mathcal{G} = (\mathcal{V}, \mathbf{A}, \mathbf{X}, \mathbf{S})$ be an attributed graph with $N$ nodes and $E$ edges, where $\mathcal{V}=\{v_1,\dots, v_N \}$ is the set of nodes.
$\mathbf{y} \in {\{0, 1\}}^N$ denotes the anomaly labels, where 1 indicates an anomalous node, and $\hat{\mathbf{y}}$ denotes the predicted labels.
The adjacency matrix is denoted as $\mathbf{A}\in {\{0,1\}}^{N\times N}$, where $\mathbf{A}_{ij}=1$ if there exists an edge between $v_i$ and $v_j$, otherwise, $\mathbf{A}_{ij}=0$.
$\mathbf{X} \in \mathbb{R}^{N\times d}$ represents the observed node attribute matrix, while
$\mathbf{S} \in \mathbb{R}^{N \times m}$ represents the sensitive attribute matrix (e.g., gender, religion, ethnicity, etc.). 
$v_i$ and $v_j$ belong to the same demographic group if $s_i=s_j$.
Here, $m$ is the total number of sensitive attributes. 
As described in \citep{deepak2020fair,sarhan2020fairness,oh2022learning}, the sensitive attribute $\mathbf{S}$ is correlated with both the observed attribute $\mathbf{X}$ and labels $\mathbf{y}$ on many real-world datasets.
The goal of fair graph anomaly detection is to provide unbiased prediction against sensitive attributes while achieving satisfactory accuracy simultaneously.
To simplify the problem, in this work, we mainly focus on a single binary sensitive attribute, i.e., $\mathbf{S}\in {\{0,1\}}^{N \times 1}$. We can easily extend our method to more complicated settings as previous studies \citep{creager2019flexibly,deepak2020fair}. More details are discussed in \autoref{sec:discussion}.

\subsection{Fairness Metrics} \label{sec:fairness_metrics}
Following \citep{agarwal2021towards,wang2022improving,zhu2023devil}, we utilize two widely used metrics to evaluate the fairness of models among demographic groups.
\textbf{Demographic Parity} \cite{dwork2012fairness} dictates the equal predicted probability across demographic groups. 
It ensures predictions are statistically unbiased to sensitive attributes, e.g., if gender is a sensitive attribute, $\Delta_{DP}$ implies that the probability of individuals from different genders being classified as anomalous should be identical.
\begin{equation}
    \Delta_{DP}=|P(\hat{y}=1|s=0) - P(\hat{y}=1|s=1)|,
\end{equation}
where $\hat{y}\in \{0,1\}$ is the predicted node label, and $\hat{y}_i=1$ indicates node $v_i$ is a predicted anomaly.

\textbf{Equal Opportunity} \citep{hardt2016equality} requires the same true positive rates of identifying anomalies for each demographic group. 
Considering gender as a sensitive attribute, $\Delta_{EO}$ encourages individuals from different genders to have an equal probability of being correctly identified as anomalous.
Given the ground-truth label $y\in \{0,1\}$ where $y_i=1$ denotes $v_i$ is a true anomaly, $\Delta_{EO}$ can be defined as:
\begin{equation}
    \Delta_{EO}=|P(\hat{y}=1|s=0, y=1) 
    - P(\hat{y}=1|s=1, y=1)| .
\end{equation}
\section{Proposed Method}
We now introduce \method, which aims to identify anomalies without skew towards demographic groups defined on sensitive attributes. 
The disentangled representation learning separates sensitive-relevant and sensitive-irrelevant representations in the latent space (see \autoref{sec:disentangle}). Fair anomaly detection is achieved through constrained reconstruction error using the disentangled representations (see \autoref{sec:fair_ad}).
We then discuss its generalization capability across diverse sensitive attributes in \autoref{sec:discussion}.
We introduce the training and inference processes in \autoref{sec:training} and provide a computational complexity analysis in \autoref{sec:complexity}.

\begin{figure*}[t]
    \centering
    \includegraphics[width=0.94\linewidth]{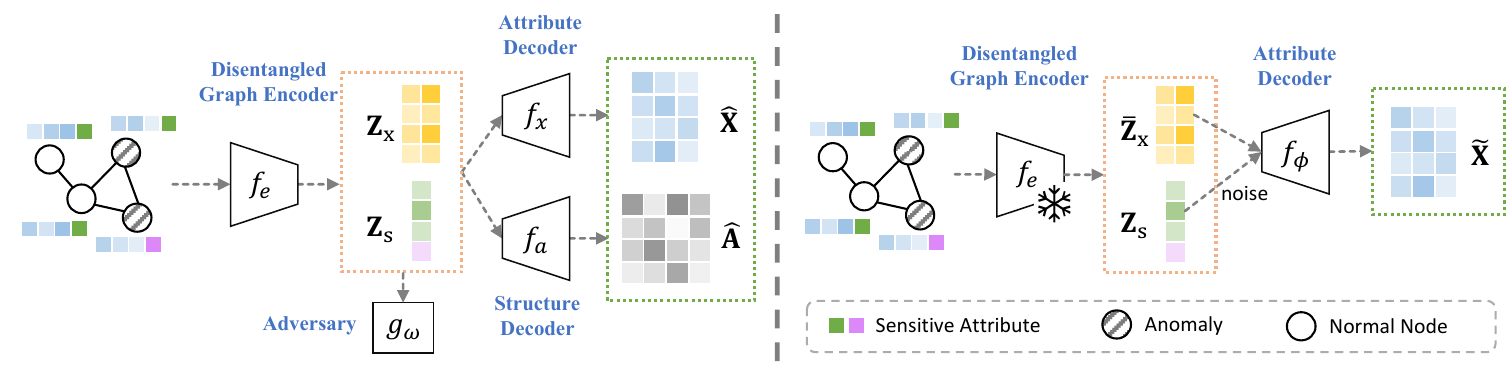}
    \caption{An overview of proposed \method framework. (Left) Disentangled fair representation learning. The disentangled graph encoder $f_e$ can separate sensitive-irrelevant representations $\mathbf{Z}_x$ and sensitive-relevant representations $\mathbf{Z}_s$ in latent space. (Right) Reconstruct-based graph anomaly detection. The constrained reconstruction error between $\mathbf{X}$ and $\tilde{\mathbf{X}}$ are used to identify anomalies. \SnowflakeChevron \ means fixing model parameters.}
    \label{fig:workflow}
\end{figure*}

\autoref{fig:workflow} illustrates the overall workflow of \method, which contains two major phases. 
Firstly, as shown in the left part of \autoref{fig:workflow}, the disentangled fair representation learning phase separates sensitive-relevant representations $\mathbf{Z}_x$ and sensitive-irrelevant representations $\mathbf{Z}_s$ in the latent space. 
Specifically, a disentangled graph encoder $f_{e}$ maps the node attribute $\mathbf{X}$ and graph topology $\mathbf{A}$ into independent sensitive-relevant and -irrelevant subspaces. Subsequently, $f_{a}$ and $f_{x}$ decode adjacency matrix $\hat{\mathbf{A}}$ and node attributes $\hat{\mathbf{X}}$, respectively. 
Ideally, $\mathbf{Z}_x$ contains no sensitive information with a well-trained $f_{e}$.
To promote the independence of $\mathbf{Z}_x$ from $\mathbf{Z}_s$, we include a learnable adversary $g_{\omega}$.
Besides minimizing reconstruction error to obtain informative representations, we also endeavor to accurately infer sensitive attributes $\mathbf{S}$ from $\mathbf{Z}_s$.
As shown in the right part of \autoref{fig:workflow}, the unsupervised graph anomaly detection phase identifies anomalies based on sensitive-irrelevant representations $\mathbf{Z}_x$.
However, the reconstruction of attributes and structures may lead to biased outcomes in decision-making, considering biases inherent in both attributes and structures. As such, the decoder $f_{\phi}$ only reconstructs node attributes from $\mathbf{Z}_x$ based on Multi-Layer Perceptron (MLP) without involving graph topology.
Moreover, as node attributes $\mathbf{X}$ inherently have potential correlations with sensitive attributes $\mathbf{S}$, we further constrain the correlation between the reconstruction error and the predicted sensitive attributes.

\subsection{Disentangled Fair Representation Learning} \label{sec:disentangle}
Informative sensitive-irrelevant representations are crucial for fair and accurate decision-making in GAD, which thoroughly considers features of nodes at both node and structural levels.
However, since the potential bias in node attributes and graph topology, GNNs might amplify sensitive information when generating node representations with the message-passing mechanism~\citep{dai2021say,zhu2023devil}.  
Previous theoretical insights and empirical evidence have highlighted the effectiveness of disentangled representation learning in separating sensitive-irrelevant representations for augmenting fairness in downstream tasks like image classification~\citep{creager2019flexibly,oh2022learning}. 
We posit that disentangled representation learning is also feasible to provide informative yet sensitive-irrelevant node representations for GAD.
The details of disentangled fair representation learning on graph-structured data in \method are described as follows.

\subsubsection{Disentangled Graph Encoder}
Based on the assumption that the latent space can be decomposed into two independent subspaces \citep{creager2019flexibly,oh2022learning}: one associated with sensitive attributes and the other devoid of them, the disentangled graph encoder $f_{e}$ strives to capture informative node representations that are irrelevant to sensitive attributes.
For graph data, the posterior distribution of node representations $q(\mathbf{Z}_x, \mathbf{Z}_s | \mathbf{X}, \mathbf{A})$ is derived from node attributes $\mathbf{X}$ and graph topology $\mathbf{A}$. 
To achieve disentanglement, it is necessary to establish conditional independence between sensitive-irrelevant representations $\mathbf{Z}_x$ and sensitive-relevant representations $\mathbf{Z}_s$, given $\mathbf{X}$ and $\mathbf{A}$.
The variational posterior distribution $q(\mathbf{Z}_x, \mathbf{Z}_s | \mathbf{X}, \mathbf{A})$ can be defined as the product of the individual distributions for $\mathbf{Z}_x$ and $\mathbf{Z}_s$:
\begin{equation}
    q(\mathbf{Z}_x, \mathbf{Z}_s | \mathbf{X}, \mathbf{A}) = q(\mathbf{Z}_x | \mathbf{X}, \mathbf{A}) q(\mathbf{Z}_s | \mathbf{X}, \mathbf{A}).
\end{equation}
As the disentanglement is conducted in latent space, 
it is crucial to capture informative node representations from both node attributes and topological structure. 
Thus, we adopt GNNs as the backbone of the disentangled graph encoder $f_{e}$.
In this work, we take Graph Convolutional Network (GCN) \citep{kipf2016semi} 
as an example. 
For the $l$-th convolutional layer, the node representation $\mathbf{H}^{(l)}$ is updated by:
\begin{equation}
    \mathbf{H}^{(l)} = \text{Conv}(\mathbf{H}^{(l-1)}, \mathbf{A}) = \phi \left( \tilde{\mathbf{D}}^{-\frac{1}{2}} \tilde{\mathbf{A}} \tilde{\mathbf{D}}^{-\frac{1}{2}} \mathbf{H}^{(l-1)} \mathbf{W}^{l} \right),
\end{equation}
where $\tilde{\mathbf{A}} = \mathbf{A} + \mathbf{I}$, $\tilde{\mathbf{D}}_{ii}=\sum_{j} \tilde{\mathbf{A}}_{ij}$, $\mathbf{I}$ is the identity matrix of $\mathbf{A}$, $\mathbf{W}^{l}$ is the weight matrix at the $l$-th layer and the initial node representation $\mathbf{H}^{(0)}$ is set to $\mathbf{X}$. 
Next, we use the reparameterization trick to estimate the sensitive-irrelevant representations $\mathbf{Z}_x$ with GCN:
\begin{equation}
    \mathbf{Z}_x = \boldsymbol{\mu} + \boldsymbol{\sigma} \cdot \bm{\varepsilon}, \ \bm{\varepsilon} \sim \mathcal{N}(\mathcal{\mathbf{0}, \mathbf{I}}),
\end{equation}
where $\mathcal{N}(\mathcal{\mathbf{0}, \mathbf{I}})$ is the standard Gaussian distribution with a mean vector of zeros $\mathbf{0}$ and an identity covariance matrix $\mathbf{I}$. The mean matrix $ \boldsymbol{\mu} = \text{Conv}_{\boldsymbol{\mu} }(\text{Conv}_{\text{shr} }(\mathbf{X}, \mathbf{A}), \mathbf{A})$ 
and the log standard deviation matrix $\log \boldsymbol{\sigma} =  \text{Conv}_{\boldsymbol{\sigma} }(\text{Conv}_{\text{shr} }(\mathbf{X}, \mathbf{A}), \mathbf{A})$ are obtained by GCNs with a shared convolutional layer $\text{Conv}_{\text{shr}}$.
Following \citep{creager2019flexibly}, we estimate the sensitive-relevant representations $ \mathbf{Z}_s $ in a deterministic manner:
\begin{equation}
    \mathbf{Z}_s = \text{Conv}_{\boldsymbol{\varphi} }(\text{Conv}_{\text{shr} }(\mathbf{X}, \mathbf{A}), \mathbf{A}).
\end{equation}
where $\text{Conv}_{\boldsymbol{\varphi}}$ is an additional layer for sensitive representations.

\subsubsection{Decoders}
In the decoding phase of disentangled representation learning, we reconstruct the attribute matrix $\mathbf{X}$, the adjacency matrix $\mathbf{A}$, and sensitive attributes $\mathbf{S}$.
Due to sensitive-irrelevant representations $\mathbf{Z}_x$ are conditionally independent of the given sensitive attributes $\mathbf{S}$, the probability $p(\mathbf{S}|\mathbf{Z}_x,\mathbf{Z}_s)$ can be simplified to $p(\mathbf{S}|\mathbf{Z}_s)$.
Consequently, the decoding phase can be formulated by: 
\begin{equation} \label{eq:decode}
    p(\mathbf{X},\mathbf{A},\mathbf{S}|\mathbf{Z}_x,\mathbf{Z}_s) = p(\mathbf{X}|\mathbf{Z}_x,\mathbf{Z}_s) p(\mathbf{A}|\mathbf{Z}_x,\mathbf{Z}_s) p(\mathbf{S}|\mathbf{Z}_s) .
\end{equation}
We adopt an attribute decoder $f_{x}$ based on GNNs to model the distribution of reconstructed node attributes $p(\mathbf{X}|\mathbf{Z}_x,\mathbf{Z}_s)$.
Besides, we consider an inner product decoder $f_a$ for structure reconstruction:
\begin{equation}
    p(\mathbf{A} | \mathbf{Z}_x, \mathbf{Z}_s) = \text{Sigmoid}(\mathbf{Z}\mathbf{Z}^{\top}),
\end{equation}
where $\mathbf{Z}=\text{Concat}(\mathbf{Z}_x, \mathbf{Z}_s)$.
Next, to model the conditional probability distribution $p(\mathbf{S}|\mathbf{Z}_s)$, we employ a binary classifier, where sensitive attributes $\mathbf{S}$ are assumed to follow a Bernoulli distribution parameterized by the Sigmoid function applied to $\mathbf{Z}_{s}$.
The conditional distribution for sensitive attributes is formulated as:
\begin{equation}
    p(\mathbf{S}|\mathbf{Z}_s)= \text{Bernoulli} (\mathbf{S} | \text{Sigmoid}(\mathbf{Z}_{s})).
\end{equation}

\subsubsection{Adversary}
To encourage the independence between sensitive-irrelevant representations $\mathbf{Z}_x$ and sensitive-relevant representations $\mathbf{Z}_s$, it is imperative that the aggregate posterior distribution can be factorized as $q(\mathbf{Z}_x,\mathbf{Z}_s)=q(\mathbf{Z}_x)q(\mathbf{Z}_s)$. Thus, we employ the Kullback-Leibler (KL) divergence between $q(\mathbf{Z}_x,\mathbf{Z}_s)$ and $q(\mathbf{Z}_x)q(\mathbf{Z}_s)$ as the disentanglement criteria, which can be expressed as ${KL}\big[q(\mathbf{Z}_x, \mathbf{Z}_s) \,\|\, q(\mathbf{Z}_x)q(\mathbf{Z}_s)\big]$. 
Following~\citep{kim2018disentangling, sugiyama2012density,creager2019flexibly}, we adopt a binary adversary $g_{\omega}$ to encourage the disentanglement by approximating the log density ratio inherent in the KL divergence term:
\begin{equation}
\begin{aligned}
{KL}\big[q(\mathbf{Z}_x, \mathbf{Z}_s) \,\|\, q(\mathbf{Z}_x)q(\mathbf{Z}_s)\big]=\mathbb{E}_{q(\mathbf{Z}_x, \mathbf{Z}_s)} \log \frac{q(\mathbf{Z}_{x},\mathbf{Z}_{s})}{q(\mathbf{Z}_{x})q(\mathbf{Z}_{s})} \\ \approx \mathbb{E}_{q(\mathbf{Z}_x, \mathbf{Z}_s)} [\log p(\tilde{\mathbf{y}}=1|\mathbf{Z}_x, \mathbf{Z}_s) - \log p(\tilde{\mathbf{y}}=0|\mathbf{Z}_x, \mathbf{Z}_s)],
\end{aligned}
\end{equation}
where $\tilde{\mathbf{y}}=1$ denotes "true" samples from the aggregate posterior $q(\mathbf{Z}_{x}, \mathbf{Z}_{s})$ and $\tilde{\mathbf{y}}=0$ denotes "fake" samples from the product of the marginals $q(\mathbf{Z}_{x})q(\mathbf{Z}_{s})$.
Specifically, we implement an MLP as $g_{\omega}$ to predict whether the sensitive-irrelevant representation $\mathbf{z}_{x}^i$ and the sensitive-relevant representation $\mathbf{z}_{s}^j$ are from the same node.

\subsubsection{Learning Objective}
We optimize the disentangled graph encoder $f_{e}$, structure decoder $f_{a}$, and attribute decoder $f_{x}$ from three aspects, including the variational lower bound term, disentanglement term, and predictiveness term.
First, the variational lower bound term comprises a reconstruction term and a KL divergence regularization term.
The reconstruction term is defined as:
\begin{equation}
    \mathcal{L}_{rec} = \mathbb{E}_{q(\mathbf{Z}_x,\mathbf{Z}_s|\mathbf{X},\mathbf{A})} \left[ (1-\epsilon ) \log p(\mathbf{X}|\mathbf{Z}_x,\mathbf{Z}_s) + \epsilon \log p(\mathbf{A}|\mathbf{Z}_x,\mathbf{Z}_s) \right ],
\end{equation}
where the first term is the reconstruction error for node attributes and the second term is for graph topology. 
Moreover, $\epsilon = \frac{\sigma_{\mathbf{X}}}{\sigma_{\mathbf{X}}+\sigma_{\mathbf{A}}}$ is the weight coefficient for automated balancing the impact of structure and attribute reconstruction \citep{liu2022bond}, where $\sigma_{\mathbf{A}}$ and $\sigma_{\mathbf{X}}$ denotes the standard deviations of $\mathbf{A}$ and $\mathbf{X}$, respectively.
The KL divergence regularization term \citep{kingma2013auto} is employed to minimize the KL divergence between the posterior distribution $q(\mathbf{Z}_x, \mathbf{Z}_s | \mathbf{X},\mathbf{A})$ and the prior distribution $p(\mathbf{Z}_x, \mathbf{Z}_s)$.
Thus, the variational lower bound term can be formulated as:
\begin{equation} \label{eq:vae_loss}
    \mathcal{L}_{vae} = \mathcal{L}_{rec} - KL[q(\mathbf{Z}_x,\mathbf{Z}_s | \mathbf{X}, \mathbf{A}) \parallel p(\mathbf{Z}_x,\mathbf{Z}_s)],
\end{equation}
where $p(\mathbf{Z}_x,\mathbf{Z}_s) = p(\mathbf{Z}_x) p(\mathbf{Z}_s)$ under the assumption that $\mathbf{Z}_x$ and $\mathbf{Z}_s$ are independent. 
The prior distributions $p(\mathbf{Z}_x)$ and $p(\mathbf{Z}_s)$ are modeled by the standard Gaussian distribution and uniform distribution, respectively.
Next, the disentanglement term, which encourages the separation of sensitive-relevant and -irrelevant representations as detailed in Eq. (10), is calculated as follows:
\begin{equation} \label{eq:dis_loss}
    \mathcal{L}_{dis} = \mathbb{E}_{\mathbf{z}_{x}^i,\mathbf{z}_{s}^i \sim q(\mathbf{Z}_x,\mathbf{Z}_s)}\log p(\tilde{y}=1|\mathbf{z}_{x}^i,\mathbf{z}_{s}^i) - \log p(\tilde{y}=0|\mathbf{z}_{x}^i,\mathbf{z}_{s}^i).
\end{equation}
Intuitively, accurate prediction of sensitive attributes can enhance the comprehensive understanding of node attributes in latent space and facilitate a clearer distinction between representations associated with sensitive attributes and those that are not.
Thus, \method incorporates a predictiveness term to align the sensitive representation $\mathbf{Z}_s$ closely with the given sensitive attributes $\mathbf{S}$:
\begin{equation}  \label{eq:clf_loss}
    \mathcal{L}_{pre} = \mathbb{E}_{q(\mathbf{Z}_s|\mathbf{X},\mathbf{A})} \log p(\mathbf{S}|\mathbf{Z}_s) .
\end{equation}
The overall loss for optimizing $f_{e}$, $f_{a}$ and $f_{x}$ can be defined as: 
\begin{equation} \label{eq:total_loss}
    \mathcal{L}_{total} = \mathcal{L}_{vae} + \gamma \mathcal{L}_{dis} + \alpha \mathcal{L}_{pre},
\end{equation}
where $\alpha$ and $\gamma$ are the weight coefficients to control the impact of the predictiveness and disentanglement terms relative to the variational lower bound term, respectively.

To train the binary adversary $g_{\omega}$,
the true sample $(\mathbf{z}_{x}^i,\mathbf{z}_{s}^i)$ is sampled from the aggregate posterior $q(\mathbf{Z}_x,\mathbf{Z}_s)$ while the fake sample $(\mathbf{z}_{x}^j,\mathbf{z}_{s}^k)$ is  sampled from the product of marginal posterior distributions  $q(\mathbf{Z}_x)q(\mathbf{Z}_{s})$.
The adversarial loss is formulated as:
\begin{equation} \label{eq:adv_loss}
    \mathcal{L}_{adv} = \mathbb{E}_{\mathbf{z}_{x}^i,\mathbf{z}_{s}^i \sim q(\mathbf{Z}_x,\mathbf{Z}_s)}\log p(\tilde{y}=1|\mathbf{z}_{x}^i,\mathbf{z}_{s}^i) 
    + \mathbb{E}_{\mathbf{z}_{x}^j,\mathbf{z}_{s}^k \sim q(\mathbf{Z}_x)q(\mathbf{Z}_{s})}\log [ 1 - p(\tilde{y}=0|\mathbf{z}_{x}^j,\mathbf{z}_{s}^k)]  .
\end{equation}
where $\tilde{y}=1$ denotes true samples and $\tilde{y}=0$ denotes fake samples.

The optimizations of $f_{e}$, $f_{a}$ and $f_{x}$ using $\mathcal{L}_{total}$ and $g_{\omega}$ using $\mathcal{L}_{adv}$ are conducted adversarially.
A well-trained disentangled graph encoder $f_{e}$, which effectively separates sensitive-irrelevant representations from node attributes and graph topology, can be adeptly employed in downstream tasks for enhancing fairness.

\subsection{Graph Anomaly Detection} \label{sec:fair_ad}
Next, the primary objective is to detect anomalies unbiased to any demographic groups based on deterministic sensitive-irrelevant representations $\bar{\mathbf{Z}}_x = \boldsymbol{\mu}$.
Since anomalies typically significantly deviate from the majority of nodes, reconstruction error has been widely used to measure anomaly scores~\citep{ding2019deep,fan2020anomalydae,shekhar2021fairod}. 
Considering the potential bias in graph topology as demonstrated in previous studies \citep{rahman2019fairwalk,spinelli2021fairdrop,zhu2023devil}, we solely reconstruct node attributes and employ MLP as the backbone of the attribute decoder $f_{\phi}$ to mitigate the impact of biased topology during the message-passing process.
The anomaly score ${o}_i$ is evaluated by reconstructing node attributes $\mathbf{X}$ from sensitive-irrelevant representations $\bar{\mathbf{Z}}_x$:
\begin{equation} \label{eq:a_score}
    {o}_i = {||\mathbf{x}_i - \tilde{\mathbf{x}}_i ||}_{F}^{2},
\end{equation}
where $\tilde{\mathbf{X}} = f_{\phi}(\bar{\mathbf{Z}}_x, \tilde{\mathbf{Z}}_{s})$ and $\tilde{\mathbf{Z}}_{s}$ denotes a shuffled variant of $\mathbf{Z}_{s}$. 
Additionally, since sensitive attributes are correlated with observed node attributes \citep{deepak2020fair,sarhan2020fairness,oh2022learning} and $\bar{\mathbf{Z}}_x$ is supposed to be devoid of sensitive information, the reconstruction error inevitably correlates with sensitive attributes.
To mitigate the impact of this correlation and prevent directly leveraging sensitive attributes, we propose a correlation constraint term, which measures the absolute correlation between the reconstruction error ${o}_i$ and predicted sensitive attributes $\mathbf{z}_{s}^{i}$:
\begin{equation} \label{eq:corr_loss}
    \mathcal{L}_{corr} = \left|\frac
    {(\sum_{i\in \mathcal{V}}{{o}_i - \mu _{o}})
    (\sum_{i\in \mathcal{V}}{\mathbf{z}_{s}^{i} - \mu _{zs}}))}
    {\sigma _{o} \sigma _{zs}} \right|,
\end{equation}
where $\mu_{o}$ and $\mu_{zs}$ are the corresponding means, while $\sigma_{o}$ and $\sigma_{zs}$ are the standard deviations of $\mathbf{o}$ and $\mathbf{Z}_s$. 

\subsubsection{Learning Objective}
In the anomaly detection phase, the encoder $f_{e}$ is set to be non-trainable, as it has already mastered the separation of sensitive-relevant and -irrelevant representations during the disentangled representation learning phase.
Accordingly, the optimization is exclusively concentrated on the decoder $f_{\phi}$.
The overall loss consisting of the reconstruction term and the correlation constraints term can be formulated as:
\begin{equation} \label{eq:ad_loss}
    \mathcal{L}_{ad} = \mathcal{L}_{rec}^{X} + \beta \mathcal{L}_{corr},
\end{equation}
where $\mathcal{L}_{rec}^{X} = \sum_{i}^{n} {o}_i$ is the attribute reconstruction loss and $\beta$ is the penalty of sensitive information in $\mathbf{X}$.

\subsection{Discussion}
\label{sec:discussion}
We analyze the generalization of \method on various types of sensitive attributes. 
\textbf{(1) Multiple Binary Sensitive Attributes.}
Let $\mathbf{s}=\left[{s^0}, {s^1}, \ldots, {s^{m-1}}\right]$ denotes the sensitive attributes for each node, where $s^j \in \{0,1\} (0 \leq j < m)$. 
The disentanglement between sensitive-irrelevant representations $\mathbf{Z}_x$ and sensitive-relevant representations $\mathbf{Z}_s \in \mathbb{R}^{N\times m}$ requires that $\mathbf{Z}_x$ is independent of each sensitive attribute $\mathbf{Z}_s^{(j)}$. 
Thus, $\mathbf{Z}_x$ and $\mathbf{Z}_s$ are disentangled if the aggregate posterior distribution can be factorized as $p(\mathbf{Z}_x,\mathbf{Z}_s) = p(\mathbf{Z}_x) \prod_j p(\mathbf{Z}_s^{(j)})$.
\textbf{(2) Single Categorical/Continuous Sensitive Attributes.}
With single binary sensitive attributes, the conditional distribution \( p(\mathbf{S}|\mathbf{Z}_s) \) is typically modeled by a Bernoulli distribution. 
If with categorical attributes, it shifts to a Multinomial distribution, i.e.
$ p(\mathbf{S}|\mathbf{Z}_s) = \text{Multinomial}(\mathbf{S}|\text{SoftMax}(\mathbf{Z}_s)).$
Similarly, for continuous attributes, \( p(\mathbf{S}|\mathbf{Z}_s) \) can be modeled by a Gaussian distribution.
\section{Experiments}
\subsection{Experimental Setup}
\textbf{Datasets.} 
We employ three real-world datasets for fair GAD, which provide both real sensitive attributes and ground-truth labels for GAD. 
In Reddit and Twitter \citep{neo2024towards} datasets, the sensitive attribute is the political leaning of users, while the anomaly label is assigned to misinformation spreaders. 
The Credit \citep{agarwal2021towards} dataset focuses on payment default detection, with age as the sensitive attribute.
Details of these datasets are summarized in \autoref{tab:dataset}. 
More details are introduced in \autoref{app:datasets}.

\begin{table}[ht]

\centering
\caption{Statistics of datasets. $\rho_G$ denotes the ratio of the minority and majority group and $\rho_A$ denotes the ratio of the anomalies and normal nodes.}
\resizebox{0.85\linewidth}{!}{%
\begin{tabular}{@{}lccccccc@{}}
\toprule
Dataset          & \# Nodes & \# Edges  & \# Attributes & $\rho_G$ & $\rho_A$ & Sensitive Attributes & Anomaly Labels \\ \midrule
\textbf{Reddit}  & 9,892    & 1,211,748 & 385           & 0.1502     & 0.1584   & Political leaning & Misinformation spreader   \\
\textbf{Twitter} & 47,712   & 468,697   & 780           & 0.1365     & 0.0713   & Political leaning & Misinformation spreader   \\ 
\textbf{Credit} & 30,000 & 1,436,858 & 13 & 0.0983 & 0.2840 & Age & Payment default \\
\bottomrule
\end{tabular}%
}
\label{tab:dataset}
\end{table}

\textbf{Baselines.}
We compare \method with  
(1) GAD methods, including \textbf{DOMINANT} \citep{ding2019deep}, \textbf{CoLA}~\citep{liu2021anomaly}, \textbf{CONAD}~\citep{xu2022contrastive}, and \textbf{VGOD}~\citep{huang2023unsupervised};
(2) GAD methods augmented with Fairness Regularizers like \textbf{FairOD} \citep{shekhar2021fairod}, \textbf{HIN} \citep{zeng2021fair}, and \textbf{Correlation} \citep{shekhar2021fairod}, which incorporate fairness constraints into optimization process; 
(3) GAD methods operated on graphs pre-processed by Graph Debiasers, such as \textbf{FairWalk} \citep{rahman2019fairwalk} and \textbf{EDITS} \citep{dong2022edits}. 
The details of baselines are introduced in \autoref{app:baselines}.

\textbf{Evaluation metrics.} 
Following ~\citep{ding2019deep,liu2021anomaly,xu2022contrastive,chai2022can,neo2024towards}, we evaluate the anomaly detection performance with the Area Under the Receiver Operating Characteristic Curve (\textbf{AUC-ROC}) and the Area Under the Precision-Recall Curve (\textbf{AUC-PR}). 
Higher AUC-ROC and AUC-PR indicate superior anomaly detection capabilities.
Regarding demographic fairness, we adopt {Demographic Parity} ($\boldsymbol{\Delta_{DP}}$) and {Equal Opportunity} ($\boldsymbol{\Delta_{EO}}$) following \citep{dai2021say,dong2022edits}. The concepts are detailed in \autoref{sec:fairness_metrics}.  Lower $\Delta_{DP}$ and $\Delta_{EO}$ suggest better fairness.
To quantitatively exhibit the performance trade-offs, we introduce \textbf{Average Rank} of two accuracy metrics and two fairness metrics \citep{wang2022improving,guo2023towards} and \textbf{Composite Score} that reflects the combined performance improvement relative to our proposed \method across all metrics.

\textbf{Implementation details.}
For \method, we use 2-layer GCN for the disentangled graph encoder $f_{e}$ and attribute decoder $f_{x}$, and 2-layer MLP for the decoder $f_{\phi}$ in the anomaly detection phase.  
The hidden dimension of each layer is fixed to be 64. 
We employ the Adam optimizer with a learning rate set to 0.001 for Reddit and Credit datasets and 0.005 for Twitter.
We set the maximum training epoch in disentangled representation learning as 100, and adopt an early stopping strategy when the loss does not decrease for 20 epochs.
In the anomaly detection phase, we train the decoder $f_{\phi}$ for 100 epochs. 
We tune $\alpha$ and $\gamma$ in~\autoref{eq:total_loss} from $\{ 0.1, 0.5, 1.0, 1.5, 2.0, 2.5\}$, and the weight of correlation constrains $\beta$ in~\autoref{eq:ad_loss} from $\{ 1e-15, 5e-15, 1e-10, 5e-10, 1e-9 \}$, respectively.
We conduct all experiments on one Linux server with an NVIDIA TESLA A800 GPU (80 GB RAM).
We run all methods ten times and report the average results to prevent extreme cases. 
More implementation details of baselines are described in  \autoref{app:implementation}.

\begin{table*}[t]
\centering
\caption{Comparison results of \method against all baseline methods on Twitter. $\uparrow$ denotes larger values are better, whereas $\downarrow$ denotes lower values are preferable. The best and second best performances are highlighted in \textbf{bold} and \underline{underlined}, respectively.}
\resizebox{0.78\textwidth}{!}{%
\begin{tabular}{@{}ll|cccc|cc@{}}
\toprule
                             &            & AUC-ROC $\uparrow$                                          & AUC-PR $\uparrow$                                           & $\Delta_{DP} \downarrow$                                  & $\Delta_{EO} \downarrow$                                                        & Comp.                                      & \multicolumn{1}{c}{Avg. Rank} \\ \midrule
\multirow{4}{*}{-}           & DOMINANT   & 56.49$\pm$0.61        & 8.92$\pm$0.15       & 4.33$\pm$0.47       & 4.51$\pm$0.44   & -29.36  & 13.25  \\
                             & CoLA       & 43.74$\pm$1.08        & 5.23$\pm$0.15       & 2.63$\pm$0.32       & 2.47$\pm$1.08   & -42.06  & 18.375 \\
                             & CONAD      & 56.12$\pm$0.73        & 8.83$\pm$0.17       & 4.08$\pm$0.47       & 4.48$\pm$0.35   & -29.54  & 13.25  \\
                             & VGOD       & 73.59$\pm$0.31        & 16.02$\pm$0.72      & 12.56$\pm$1.01      & 11.49$\pm$1.84  & -20.37  & 14     \\ \midrule
\multirow{4}{*}{FairWalk}    & DOMINANT   & 53.06$\pm$0.81        & 8.17$\pm$0.43       & 1.32$\pm$0.27       & 1.27$\pm$0.35   & -27.29  & 11.75  \\
                             & CoLA       & 49.02$\pm$0.54        & 6.34$\pm$0.23       & 0.23$\pm$0.14       & 0.34$\pm$0.16   & -31.14  & 10.75  \\
                             & CONAD      & 53.36$\pm$0.70        & 8.32$\pm$0.42       & 1.30$\pm$0.18       & 1.37$\pm$0.39   & -26.92  & 11     \\
                             & VGOD       & 60.21$\pm$0.26        & 9.14$\pm$0.15       & 9.84$\pm$0.27       & 4.98$\pm$0.37   & -31.40  & 14.25  \\ \midrule
\multirow{4}{*}{EDITS}       & DOMINANT   & 53.79$\pm$0.25        & 8.75$\pm$0.06       & 2.63$\pm$0.05       & 2.21$\pm$0.15   & -28.23  & 12.75  \\
                             & CoLA       & 46.62$\pm$1.35        & 5.70$\pm$0.30       & 1.84$\pm$0.39       & 1.03$\pm$0.82   & -36.48  & 14.25  \\
                             & CONAD      & 53.83$\pm$0.24        & 8.75$\pm$0.06       & 2.62$\pm$0.03       & 2.16$\pm$0.13   & -28.13  & 11.875 \\
                             & VGOD       & 82.06$\pm$0.87        & 25.75$\pm$1.40      & 20.43$\pm$1.42      & 22.32$\pm$2.45  & -20.87  & 13     \\ \midrule
\multirow{4}{*}{FairOD}      & DOMINANT & 53.69$\pm$3.63 & 7.87$\pm$1.10 & 2.56$\pm$1.51 & {2.49$\pm$1.52} & -29.42 & 14.25 \\
                             & CoLA & 45.81$\pm$7.40 & 6.29$\pm$1.52 & 1.95$\pm$1.82 & {1.23$\pm$0.77} & -37.01 & 14.625 \\
& CONAD & 57.09$\pm$0.43 & 9.08$\pm$0.13 & 4.48$\pm$0.45 & {4.59$\pm$0.56} & -28.83 & 13.75 \\
& VGOD & 76.27$\pm$1.12 & 15.56$\pm$0.54 & 9.75$\pm$1.07 & {5.61$\pm$1.67} & -9.46 & 12.5 \\ \midrule
\multirow{4}{*}{HIN} & DOMINANT & 54.07$\pm$2.22 & 8.23$\pm$0.79 & 2.50$\pm$1.24 & {3.51$\pm$0.96} & -29.64 & 13 \\
& CoLA & 48.08$\pm$5.31 & 6.49$\pm$1.18 & 2.38$\pm$1.23 & {0.62$\pm$0.69} & -34.36 & 12.75 \\
& CONAD & 53.74$\pm$2.74 & 8.12$\pm$1.02 & 2.59$\pm$1.36 & {3.34$\pm$1.04} & -30.00 & 14.25 \\
& VGOD & {81.58$\pm$1.88} & {18.86$\pm$1.00} & 12.52$\pm$0.91 & {9.32$\pm$1.38} & \underline{-7.33} & 12.5 \\ \midrule
\multirow{4}{*}{Correlation} & DOMINANT & 56.21$\pm$0.60 & 8.81$\pm$0.14 & 4.22$\pm$0.38 & {4.58$\pm$0.29} & -29.71 & 14.25 \\
& CoLA & 48.65$\pm$2.89 & 6.29$\pm$0.48 & 3.32$\pm$1.18 & {3.42$\pm$2.10} & -37.73 & 17.875 \\
& CONAD & 55.94$\pm$0.71 & 8.76$\pm$0.17 & 4.00$\pm$0.44 & {4.52$\pm$0.31} & -29.75 & 14.25 \\
& VGOD & 71.78$\pm$0.58 & 11.80$\pm$0.24 & 5.22$\pm$0.56 & {{0.76$\pm$0.53}} & -8.33 & \underline{8.75} \\ \midrule
Ours & \method & 75.59$\pm$2.00 & 11.85$\pm$0.79 & {0.72$\pm$0.70} & {0.79$\pm$0.61} & \textbf{0} & \textbf{3.75} \\ \bottomrule
\end{tabular}%
}
\label{tab:overall_twitter}
\end{table*}

\begin{table*}[t]
\centering
\caption{Comparison results of \method against all baseline methods on Reddit. $\uparrow$ denotes larger values are better, whereas $\downarrow$ denotes lower values are preferable. The best and second best performances are highlighted in \textbf{bold} and \underline{underlined}, respectively. OOM indicates out of memory.}
\resizebox{0.8\textwidth}{!}{%
\begin{tabular}{@{}ll|cccc|cc@{}}
\toprule
& & AUC-ROC $\uparrow$ & AUC-PR $\uparrow$ & $\Delta_{DP} \downarrow$ & $\Delta_{EO} \downarrow$ & Comp. & Avg. Rank \\ \midrule
\multirow{4}{*}{-} & DOMINANT & 60.82$\pm$0.09 & 20.02$\pm$0.04 & 13.20$\pm$0.08 & 5.59$\pm$0.19 & -13.01 & 10.25 \\
& CoLA & 45.20$\pm$0.98 & 17.90$\pm$1.54 & 4.95$\pm$1.78 & 4.06$\pm$1.85 & -20.97 & 12.5 \\
& CONAD & 60.81$\pm$0.10 & 20.02$\pm$0.05 & 13.32$\pm$0.33 & 5.70$\pm$0.37 & -13.25 & 11.5 \\
& VGOD & 72.01$\pm$0.93 & {39.38$\pm$2.45} & 42.47$\pm$5.61 & 46.79$\pm$6.09 & -52.93 & 12 \\ \midrule
\multirow{4}{*}{FairWalk} & DOMINANT & 51.26$\pm$1.08 & 14.55$\pm$0.65 & 2.45$\pm$0.79 & 1.97$\pm$1.25 & -13.67 & 11 \\
& CoLA & 51.12$\pm$0.83 & 14.83$\pm$0.85 & {0.69$\pm$0.37} & {0.60$\pm$0.59} & \underline{-10.40} & 10 \\
& CONAD & 51.26$\pm$1.88 & 14.68$\pm$1.15 & 2.49$\pm$1.28 & 2.52$\pm$1.58 & -14.13 & 11.25 \\
& VGOD & 67.08$\pm$0.61 & 28.53$\pm$0.28 & 31.99$\pm$0.53 & 29.89$\pm$1.02 & -41.33 & 11.75 \\ \midrule
\multirow{4}{*}{EDITS} & DOMINANT & OOM & OOM & OOM & OOM & - & - \\
& CoLA & 54.41$\pm$1.62 & 23.13$\pm$3.67 & 23.74$\pm$5.76 & 21.00$\pm$5.38 & -42.26 & 14 \\
& CONAD & OOM & OOM & OOM & OOM & - & - \\
& VGOD & OOM & OOM & OOM & OOM & - & - \\ \midrule
\multirow{4}{*}{FairOD} & DOMINANT & 60.94$\pm$0.07 & 19.97$\pm$0.08 & 13.08$\pm$0.21 & 5.04$\pm$0.15 & -12.27 & \underline{9.5} \\
& CoLA & 45.76$\pm$10.23 & 15.77$\pm$5.42 & 13.26$\pm$6.24 & 5.98$\pm$5.83 & -32.77 & 16.5 \\
& CONAD & 60.53$\pm$0.11 & 19.53$\pm$0.08 & 12.57$\pm$0.11 & 5.02$\pm$0.11 & -12.59 & 11 \\
& VGOD & 71.65$\pm$2.55 & 30.40$\pm$4.98 & 28.66$\pm$7.70 & 37.74$\pm$4.28 & -39.41 & 11.5 \\ \midrule
\multirow{4}{*}{HIN} & DOMINANT & 60.91$\pm$0.06 & 20.10$\pm$0.04 & 13.38$\pm$0.10 & 5.69$\pm$0.22 & -13.12 & 10.5 \\
& CoLA & 45.52$\pm$1.47 & 18.59$\pm$1.68 & 4.58$\pm$2.50 & 4.43$\pm$2.28 & -19.96 & 12.25 \\
& CONAD & 60.89$\pm$0.18 & 20.06$\pm$0.10 & 13.35$\pm$0.21 & 5.81$\pm$0.45 & -13.27 & 11.25 \\
& VGOD & {72.66$\pm$3.32} & 26.18$\pm$3.36 & 17.92$\pm$6.12 & 30.39$\pm$9.38 & -24.53 & 10.75 \\ \midrule
\multirow{4}{*}{Correlation} & DOMINANT & 60.38$\pm$0.08 & 19.57$\pm$0.06 & 12.27$\pm$0.25 & 4.42$\pm$0.20 & -11.80 & 10.25 \\
& CoLA & 50.94$\pm$6.37 & 15.29$\pm$3.50 & 9.63$\pm$7.26 & 11.68$\pm$8.23 & -30.14 & 15.25 \\
& CONAD & 59.51$\pm$0.33 & 18.86$\pm$0.28 & 10.94$\pm$0.74 & 3.80$\pm$0.35 & -11.43 & 10.25 \\
& VGOD & {75.68$\pm$3.07} & {31.09$\pm$2.83} & 32.42$\pm$9.53 & 42.72$\pm$9.65 & -43.43 & 11.25 \\ \midrule
Ours & \method & 60.67$\pm$0.42 & 16.46$\pm$0.25 & {0.94$\pm$0.72} & {1.13$\pm$0.81} & \textbf{0} & \textbf{8} \\ \bottomrule
\end{tabular}%
}
\label{tab:overall_reddit}
\end{table*}

\begin{table*}[t]
\centering
\caption{Comparison results of \method against all baseline methods on Credit. $\uparrow$ denotes larger values are better, whereas $\downarrow$ denotes lower values are preferable. The best and second best performances are highlighted in \textbf{bold} and \underline{underlined}, respectively.}
\resizebox{0.8\textwidth}{!}{%
\begin{tabular}{@{}ll|cccc|cc@{}}
\toprule
                             &                               & \multicolumn{1}{c}{AUC-ROC $\uparrow$} & \multicolumn{1}{c}{AUC-PR $\uparrow$} & \multicolumn{1}{c}{$\Delta_{DP} \downarrow$} & \multicolumn{1}{c}{$\Delta_{EO} \downarrow$} & \multicolumn{1}{c}{Comp.} & \multicolumn{1}{c}{Avg. Rank} \\ \midrule
\multirow{4}{*}{-}           & \multicolumn{1}{l|}{DOMINANT} & 52.34$\pm$4.14                          & 25.28$\pm$4.33                        & 1.63$\pm$1.00                                & \multicolumn{1}{l|}{2.34$\pm$1.62}           & -2.11                   & 8.75                           \\
                             & \multicolumn{1}{l|}{CoLA}     & 46.31$\pm$0.94                          & 18.96$\pm$0.54                         & 3.04$\pm$1.22                                & \multicolumn{1}{l|}{1.48$\pm$1.26}           & -15.01                   & 16                             \\
                             & \multicolumn{1}{l|}{CONAD}    & 53.68$\pm$5.28                         & 26.02$\pm$5.03                        & 3.91$\pm$2.77                                & \multicolumn{1}{l|}{4.19$\pm$2.61}           & -4.16                   & 10.75                          \\
                             & \multicolumn{1}{l|}{VGOD}     & 54.81$\pm$0.90                          & 26.02$\pm$0.86                         & 17.54$\pm$1.83                               & \multicolumn{1}{l|}{13.59$\pm$1.84}          & -26.06                   & 14.75                          \\ \midrule
\multirow{4}{*}{FairWalk}    & \multicolumn{1}{l|}{DOMINANT} & 49.56$\pm$0.40                          & 21.78$\pm$0.44                         & 0.67$\pm$0.74                                  & \multicolumn{1}{l|}{1.50$\pm$0.99}            & -6.60                   & 11.75                          \\
                             & \multicolumn{1}{l|}{CoLA}     & 50.29$\pm$0.17                          & 22.63$\pm$0.31                         & 0.32$\pm$0.28                                  & \multicolumn{1}{l|}{0.22$\pm$0.23}             & -6.32                   & 8                              \\
                             & \multicolumn{1}{l|}{CONAD}    & 49.76$\pm$0.55                          & 21.91$\pm$0.63                         & 1.21$\pm$0.83                                 & \multicolumn{1}{l|}{1.02$\pm$0.80}            & -3.38                   & 10.25                          \\
                             & \multicolumn{1}{l|}{VGOD}     & 49.33$\pm$0.10                          & 22.00$\pm$0.06                         & 31.57$\pm$0.94                                & \multicolumn{1}{l|}{29.43$\pm$1.17}          & -65.44                  & 22                             \\ \midrule
\multirow{4}{*}{EDITS}       & \multicolumn{1}{l|}{DOMINANT} & 49.19$\pm$1.71                         & 22.06$\pm$1.85                        & 2.99$\pm$2.12                                & \multicolumn{1}{l|}{2.73$\pm$1.59}           & -10.23                   & 14.5                           \\
                             & \multicolumn{1}{l|}{CoLA}     & 49.82$\pm$1.01                         & 21.81$\pm$0.73                         & 5.07$\pm$2.79                                & \multicolumn{1}{l|}{2.95$\pm$2.40}           & -13.54                   & 16.25                          \\
                             & \multicolumn{1}{l|}{CONAD}    & 49.60$\pm$1.50                         & 22.36$\pm$1.14                        & 5.23$\pm$4.18                                & \multicolumn{1}{l|}{4.51$\pm$2.66}           & -12.14                   & 17                             \\
                             & \multicolumn{1}{l|}{VGOD}     & 46.61$\pm$3.23                         & 20.52$\pm$1.61                        & 9.32$\pm$10.38                               & \multicolumn{1}{l|}{9.27$\pm$9.20}           & -27.23                   & 21.75                          \\ \midrule
\multirow{4}{*}{FairOD}      & \multicolumn{1}{l|}{DOMINANT} & 52.59$\pm$4.01                         & 24.64$\pm$4.29                        & 4.37$\pm$3.54                                & \multicolumn{1}{l|}{5.04$\pm$3.67}           & -7.95                   & 13.5                           \\
                             & \multicolumn{1}{l|}{CoLA}     & 47.43$\pm$6.33                         & 21.24$\pm$4.40                        & 7.99$\pm$5.79                                & \multicolumn{1}{l|}{6.09$\pm$4.93}           & -21.18                   & 20.25                          \\
                             & \multicolumn{1}{l|}{CONAD}    & 55.03$\pm$5.39                         & 27.38$\pm$5.14                        & 4.10$\pm$3.59                                & \multicolumn{1}{l|}{4.02$\pm$3.66}           & -1.48                   & 8.5                            \\
                             & \multicolumn{1}{l|}{VGOD}     & 55.70$\pm$1.32                         & 26.87$\pm$1.39                        & 16.76$\pm$1.29                               & \multicolumn{1}{l|}{13.02$\pm$1.35}          & -22.98                   & 13                             \\ \midrule
\multirow{4}{*}{HIN}         & \multicolumn{1}{l|}{DOMINANT} & 51.98$\pm$4.27                         & 24.88$\pm$4.38                        & 1.81$\pm$1.28                                & \multicolumn{1}{l|}{2.40$\pm$1.60}           & -3.11                   & 10                             \\
                             & \multicolumn{1}{l|}{CoLA}     & 46.31$\pm$0.94                          & 18.96$\pm$0.54                         & 3.06$\pm$1.20                                & \multicolumn{1}{l|}{1.45$\pm$1.28}           & -15.01                   & 15.5                           \\
                             & \multicolumn{1}{l|}{CONAD}    & 53.77$\pm$3.69                         & 25.40$\pm$3.54                        & 2.73$\pm$1.48                                & \multicolumn{1}{l|}{2.66$\pm$1.71}           & -1.99                   & 8.25                           \\
                             & \multicolumn{1}{l|}{VGOD}     & 57.26$\pm$3.19                         & 28.15$\pm$2.66                        & 13.54$\pm$4.17                               & \multicolumn{1}{l|}{11.11$\pm$3.78}          & -15.00                   & 11.5                           \\ \midrule
\multirow{4}{*}{Correlation} & \multicolumn{1}{l|}{DOMINANT} & 52.37$\pm$4.11                         & 25.33$\pm$4.28                        & 1.55$\pm$0.99                                 & \multicolumn{1}{l|}{2.27$\pm$1.62}           & \underline{-1.87}                 & \underline{7.75}                           \\
                             & \multicolumn{1}{l|}{CoLA}     & 50.58$\pm$7.68                         & 23.33$\pm$5.34                        & 4.92$\pm$3.18                                & \multicolumn{1}{l|}{6.32$\pm$4.14}           & -13.10                   & 15.75                          \\
                             & \multicolumn{1}{l|}{CONAD}    & 53.68$\pm$5.26                         & 26.02$\pm$5.03                        & 3.87$\pm$2.78                                & \multicolumn{1}{l|}{4.13$\pm$2.55}           & -4.07                   & 10.25                          \\
                             & \multicolumn{1}{l|}{VGOD}     & 55.86$\pm$1.10                         & 26.85$\pm$0.99                         & 12.42$\pm$1.42                               & \multicolumn{1}{l|}{8.27$\pm$1.90}           & -13.75                   & 11.75                          \\ \midrule
Ours                         & \method        & 55.41$\pm$1.80                         & 23.70$\pm$0.45                         & 1.78$\pm$1.12                                & 1.57$\pm$1.00                                & \textbf{0.00}           & \textbf{7.25}                  \\ \bottomrule
\end{tabular}%
}
\label{tab:overall_credit}
\end{table*}
\begin{figure}[t]
    \centering
    \includegraphics[width=0.92\linewidth]{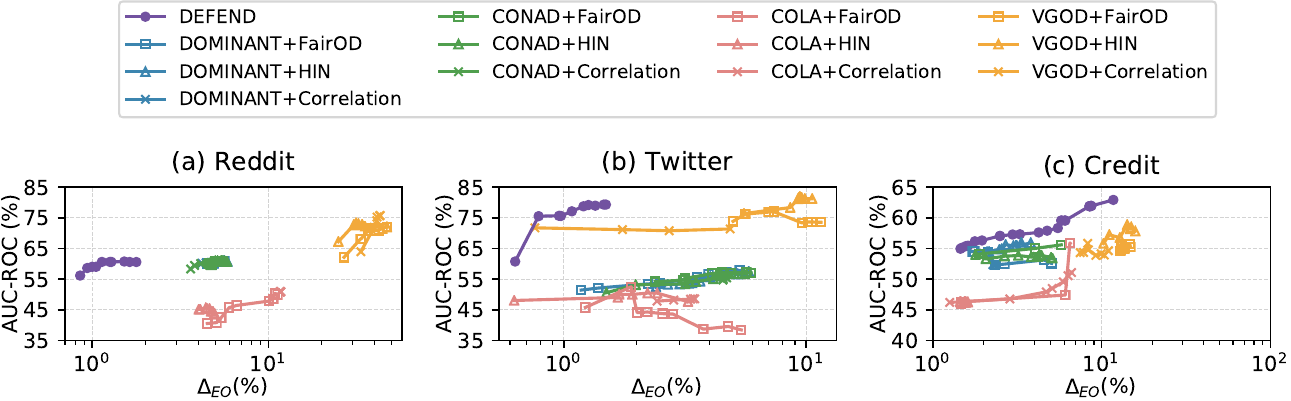}
    \caption{Fairness-Utility trade-off curves of different methods on three datasets. The upper-left corner is optimal, which has high AUC-ROC and low $\Delta_{EO}$.}
    \label{fig:tradeoff_curve}
\end{figure}

\subsection{Performance Comparison}
\subsubsection{Fairness and Utility Performance}
\autoref{tab:overall_twitter}, \autoref{tab:overall_reddit} and \autoref{tab:overall_credit} show the fairness ($\Delta_{DP}$ and $\Delta_{EO}$) and utility (AUC-ROC and AUC-PR) performance of \method and all baselines on Twitter and Reddit, respectively. 
We have the following observations:
\textbf{(1) \method demonstrates superior performance in balancing fairness and accuracy trade-offs.} To evaluate this trade-off quantitatively, we analyze the average ranks and relative improvements across two accuracy metrics and two fairness metrics. The runner-up performing model shows a 7.33\% decline in performance compared to \method, with \method achieving an average rank of 3.75 versus 8.75 for the runner-up model on Twitter. \autoref{fig:tradeoff_curve} illustrates the fairness-utility trade-off curve, which further validates the superiority of \method (detailed analysis provided in Section \ref{sec:trade-off}).
\textbf{(2) Standard GAD methods exhibit a clear trade-off between accuracy and fairness.} For instance, VGOD with Correlation achieves superior detection performance with AUC-ROC reaching 75.68\% and AUC-PR attaining 31.09\%, but demonstrates substantial fairness violations with $\Delta_{DP}$ at 32.42\% and $\Delta_{EO}$ at 42.72\% on Reddit. This observation reveals an inherent tension between detection capability and fairness in existing approaches.
\textbf{(3) The integration of graph debiasers and fairness regularizers typically improves fairness.} For example, DOMINANT with FairWalk achieves better fairness with $\Delta_{DP}$ of 2.45\% and $\Delta_{EO}$ of 1.97\%, while DOMINANT with FairOD achieves better fairness with $\Delta_{DP}$ of 13.08\% and $\Delta_{EO}$ of 5.04\% on Reddit. However, these fairness improvements sometimes comes at the cost of reduced detection performance, e.g. AUC-ROC of DOMINANT with FairWalk dropping from 60.82\% to 51.26\% on Reddit.

\subsubsection{Trade-off between Fairness and Utility} \label{sec:trade-off}
We extend our analysis to a comparative evaluation of the fairness-utility trade-off performance of \method with GAD methods incorporating fairness regularizers. 
We choose $\Delta_{EO}$ to measure fairness and AUC-ROC for utility assessment. 
Each method is trained on a range of hyperparameters and the resulting Pareto front curves are presented in \autoref{fig:tradeoff_curve}.
Notably, the optimal point is at the upper-left corner, which has perfect accuracy and fairness, reflected by a high AUC-ROC coupled with a low $\Delta_{EO}$. 
Specifically, a high AUC-ROC signified proficient detection of both normal and anomalous nodes, whereas a low  $\Delta_{EO}$ indicates an equivalent probability of correctly identifying anomalies across different demographic groups.
In contrast, the upper-right corner represents high performance at the cost of poor fairness, while the lower-left exhibits strong fairness but compromised performance.
From \autoref{fig:tradeoff_curve}, we can observe the trade-off curve of \method exhibits a superior distribution near the optimal point than the baselines.

\begin{table*}[t]
\centering
\caption{Comparison results of \method and its variants. $\uparrow$ denotes larger values are better, whereas $\downarrow$ denotes lower values are preferable. The best and second best performances are highlighted in \textbf{bold} and \underline{underlined}, respectively.}
\resizebox{0.78\textwidth}{!}{%
\begin{tabular}{@{}c|c|cccc|cc@{}}
\toprule
Dataset & Method & AUC-ROC $\uparrow$ & AUC-PR $\uparrow$ & $\Delta_{DP} \downarrow$ & $\Delta_{EO} \downarrow$ & Comp. & Avg. Rank \\ 
\midrule
\multirow{5}{*}{Reddit} 
 & \method-D  & \textbf{64.41$\pm$1.23}   & \textbf{20.50$\pm$0.82}   & 14.36$\pm$1.23                & 13.57$\pm$1.66 & -18.08 & \underline{3} \\
 & \method-C  & \underline{64.54$\pm$0.75} & \underline{20.43$\pm$0.45} & 13.95$\pm$1.79               & 12.51$\pm$3.03 & -16.55 & \underline{3} \\
 & \method-A  & 60.44$\pm$1.06            & 16.39$\pm$0.41             & 1.64$\pm$1.36                 & 2.53$\pm$1.99 & \underline{-2.4} & 3.5 \\
 & \method{}+S & 54.74$\pm$1.02           & 14.78$\pm$0.33             & \textbf{0.90$\pm$0.54}        & \underline{1.50$\pm$1.39} & -7.94 & 3.25 \\ 
 & \method    & 60.67$\pm$0.42            & 16.46$\pm$0.25             & \underline{0.94$\pm$0.72}     & \textbf{1.13$\pm$0.81} & \textbf{0.00} & \textbf{2.25} \\ 
\midrule
\multirow{5}{*}{Twitter} 
 & \method-D  & \underline{87.44$\pm$0.14} & \underline{23.63$\pm$0.19} & 15.53$\pm$0.22               & 13.85$\pm$1.20 & -4.24 & \underline{3} \\
 & \method-C  & \textbf{87.58$\pm$0.16}   & \textbf{23.85$\pm$0.27}    & 15.86$\pm$0.47                & 15.31$\pm$0.97 & -5.67 & \underline{3} \\
 & \method-A  & 83.29$\pm$4.70            & 18.83$\pm$5.34             & 8.89$\pm$7.18                 & 8.37$\pm$7.03 & \underline{-1.07} & \underline{3} \\
 & \method{}+S & 50.15$\pm$0.74           & 6.48$\pm$0.13              & \underline{1.28$\pm$0.19}     & \underline{1.59$\pm$1.21} & -32.17 & 3.5 \\ 
 & \method    & 75.59$\pm$2.00            & 11.85$\pm$0.79             & \textbf{0.72$\pm$0.70}        & \textbf{0.79$\pm$0.61} & \textbf{0.00} & \textbf{2.5} \\ 
 \midrule
\multirow{5}{*}{Credit} 
 & \method-D  & \textbf{67.60$\pm$0.19} & \textbf{36.12$\pm$0.27} & 18.98$\pm$0.42               & 22.04$\pm$0.94 & -13.06 & \underline{3} \\
 & \method-C  & \underline{67.23$\pm$1.04}   & \underline{35.83$\pm$1.13}    & 18.42$\pm$1.07                & 20.96$\pm$1.77 & -12.08 & \underline{3} \\
 & \method-A  & 54.69$\pm$1.27            & 23.47$\pm$0.36             & \underline{1.56$\pm$0.79}                 & 2.45$\pm$1.35 & \underline{-1.60} & 3.25 \\
 & \method{}+S & 49.08$\pm$1.35           & 21.65$\pm$0.75              & \textbf{1.33$\pm$1.61}     & \underline{1.66$\pm$1.25} & -8.03 & 3.25 \\ 
 & \method    & 55.41$\pm$1.80            & 23.70$\pm$0.45             & 1.78$\pm$1.12        & \textbf{1.57$\pm$1.00} & \textbf{0.00} & \textbf{2.5} \\ 
\bottomrule
\end{tabular}
}
\label{tab:ablation}
\end{table*}

\subsection{Ablation Study}
We conduct ablation studies to evaluate the effectiveness of key components of \method.
We verify the performance of four variants of \method: 
(1) \method without correlation constraints term in \autoref{eq:ad_loss} in anomaly detection (\textbf{\method-C}).
(2) \method without the disentangled representation learning (\textbf{\method-D}).
(3) \method without the adversary $g_\omega$ in disentangled representation learning (\textbf{\method-A}).
(4) \method with a dot product decoder to reconstruct the graph structure for anomaly detection (\textbf{\method{}+S}).
The results are shown in \autoref{tab:ablation}.

The following observations can be made from \autoref{tab:ablation}.
\textbf{(1)} \method outperforms all variants in fairness, which validates the importance of each component in promoting fairness for GAD. 
Nevertheless, the accuracy of \method is surpassed by several variants, such as \method-C and \method-D, likely due to the stringent fairness requirements. 
As illustrated in \autoref{fig:tradeoff_curve}~(b), relaxing the fairness constraints can improve utility performance.
Specifically, the trade-off curve shows that \method achieves an AUC-ROC of nearly 80\% with a $\Delta_{EO}$ of about 1.5\%.
In this case, \method not only exceeds its variants in fairness but also maintains a comparable utility performance to them.
\textbf{(2)} The fairness of \method-C is significantly worse than \method, indicating a notable correlation between reconstruction error and sensitive attributes.
\textbf{(3)} The fairness performance of \method-D further drops than \method, highlighting the efficacy of disentangled sensitive-irrelevant representations in enhancing fairness in downstream tasks.
For example, the $\Delta_{EO}$ of \method-D improves 1\% and 12\% than that of \method-C and \method on Reddit, respectively.
\textbf{(4)} \method-A underscores the effectiveness of the adversary in enhancing fairness, although this comes at a cost of accuracy. For instance, the improvement in AUC-ROC from 75.59\% to 83.29\% is accompanied by a considerable decrease in $\Delta_{DP}$ from 0.72\% to 8.89\%. In some cases, sacrificing partial accuracy to achieve higher fairness is a justified trade-off.
\textbf{(5)} Compared to \method{}+S, \method not only achieves superior accuracy but also demonstrates enhanced fairness in most cases, which suggests that biased graph topology, negatively impacts the performance of reconstruct-based GAD method. 
This finding supports our choice to avoid structure reconstruction in the anomaly detection phase.

\subsection{Parameter Analysis}
In this section, we investigate the impact of three key parameters within \method on Reddit, including $\alpha$ and $\gamma$ controlling the weight of prediction term and disentanglement term in disentangled representation learning, while $\beta$ controlling the weight of correlation constraints in the anomaly detection phase.

\subsubsection{Impact of \texorpdfstring{$\alpha$}{alpha} and \texorpdfstring{$\gamma$}{gamma}}
To investigate the influence of predictiveness and disentanglement terms, We train \method with the values of $\alpha$ and $\gamma$ among $\{ 0.1, 0.5, 1.0, 1.5, 2.0, 2.5, 5.0\}$. 
The results in terms of AUC-ROC and $\Delta_{EO}$ are presented in \autoref{fig:alpha_gamma_sens} (a) and \autoref{fig:alpha_gamma_sens} (b), respectively.
\autoref{fig:alpha_gamma_sens} (a) reveals that the anomaly detection efficacy remains substantially stable when $\alpha \geq 2.0$ and $\gamma \leq 1.0$, which represents an optimal range for anomaly detection capabilities. 
When $\gamma \geq 1.0$, there is a marked enhancement in fairness at the expense of a rapid decline in accuracy.
This trade-off occurs because larger $\gamma$ forces the model to achieve more similar predictions across demographic groups, potentially compromising its ability to identify true anomalies.
Besides, a reduction in $\alpha$ alone significantly decreases both AUC-ROC and $\Delta_{EO}$. 
It underscores that without accurate sensitive attribute prediction, the model fails to effectively disentangle the representations, leading to both poor detection performance and unfair predictions.
Therefore, selecting appropriate values for $\alpha$ and $\gamma$ is instrumental in navigating the tradeoff between anomaly detection accuracy and fairness.

\begin{minipage}{0.55\textwidth}
    \begin{figure}[H]
        \centering
        \subfigure[AUC-ROC (\%)]{\includegraphics[width=0.45\textwidth]{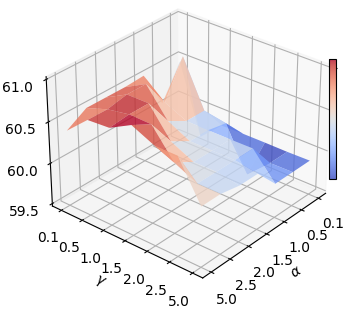}}
        \subfigure[$\Delta_{EO}$(\%)]{\includegraphics[width=0.45\textwidth]{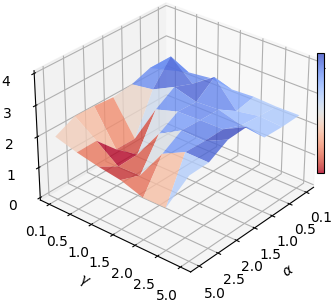}}
        \caption{Impacts of varying predictiveness term weight $\alpha$ and disentanglement term weight $\gamma$ in \autoref{eq:total_loss} on Reddit dataset in terms of AUC-ROC and $\Delta_{EO}$.}
        \label{fig:alpha_gamma_sens}
    \end{figure}
\end{minipage}
\hfill
\begin{minipage}{0.42\textwidth}
    \begin{figure}[H]
        \centering
        \includegraphics[width=0.9\textwidth]{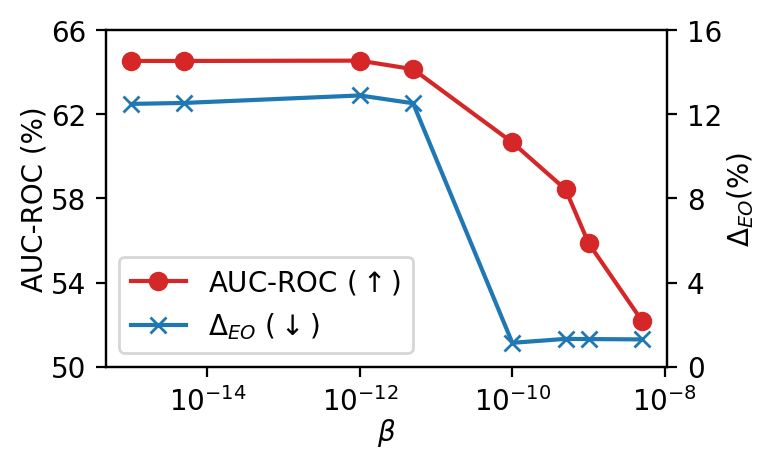}
        \vspace{0.1in}
        \caption{Impacts of varying correlation constraints weight $\beta$ in \autoref{eq:ad_loss} on Reddit.}
        \label{fig:beta_sens}
    \end{figure}
\end{minipage}

\subsubsection{Impact of \texorpdfstring{$\beta$}{beta}}
To evaluate the effect of the constraint term, we further train \method with different values of $\beta$ among $\{ 1e-15, 5e-15, 1e-12, 5e-12, 1e-10, 5e-10, 1e-9, 5e-9 \}$ on Reddit. 
The results are depicted in \autoref{fig:beta_sens}.
First, we can observe that the $\Delta_{EO}$ decreases more than 10\% while the AUC-ROC witnesses a marginal decline of nearly 4\% at $\beta$ around $5e-12$. 
This favorable trade-off demonstrates that our constraint mechanism effectively reduces the model's reliance on sensitive information during anomaly scoring while largely preserving its detection capability.
It emphasizes the necessity of integrating constraints within the reconstruction-based evaluation of anomalous nodes, considering the interrelation between input and sensitive attributes.
Moreover, it is evident that both AUC-ROC and $\Delta_{EO}$ exhibit a decrement as the weight assigned to correlation constraints increases.
It reveals that stronger constraints force the reconstruction process to be more independent of sensitive attributes, leading to fairer but potentially less discriminative representations.
The judicious selection of $\beta$ is crucial for the trade-off between utility and fairness.
\section{Related Work}
\subsection{Graph Anomaly Detection}
Graph anomaly detection (GAD) has drawn rising interest since graph-structured data becoming increasingly prevalent in complex real-world systems. 
Given the absence of ground truth anomaly labels, many GAD methods focus on an unsupervised manner.
Autoencoder is a prominent paradigm in this domain, which hypothesizes that the decoder cannot properly reconstruct anomalies deviating significantly from the majority.
DOMINANT \citep{ding2019deep} employs GCN-based autoencoder and evaluates anomalies based on the reconstruction errors of node attributes and graph topology.
AnomalyDAE \citep{fan2020anomalydae} adopts a dual autoencoder structure to capture the cross-modality interactions between topology and node attributes.
GAD-NR \citep{roy2023gad} further incorporates neighborhood reconstruction.
Contrastive learning is another prevalent self-supervised paradigm in GAD. 
CoLA \citep{liu2021anomaly} firstly introduces node-subgraph contrast to identify anomalies based on the relation between nodes and their neighbors.
Building upon this, ANEMONE \citep{jin2021anemone} incorporates node-node contrast 
and GRADATE~\citep{duan2023graph} incorporates subgraph-subgraph contrast to explore multi-level characteristics.
Several studies also investigate anomaly detection across different levels of supervision \citep{chang2024multitask, xu2024lego, liudata, liu2024tgtod}.
Despite the efficacy of these methods in anomaly detection, they are prone to biased decisions due to the neglect of sensitive attributes.
To bridge this gap, we explore a novel problem of fair graph anomaly detection, aiming to detect anomalies impartially, without bias toward sensitive attributes.

\subsection{Fairness on Graphs}
Numerous studies have been conducted to mitigate source bias in training data to promote fairness in decision-making for graph learning tasks~\citep{dong2023fairness,chen2023fairness}. 
Graph debiasing methods involve removing bias from the input graph before conducting target tasks. 
For example, 
FairWalk \citep{rahman2019fairwalk} enhances the general random walk algorithm to capture more diverse neighborhoods, thereby producing embeddings that exhibit reduced bias,
while FairDrop \citep{spinelli2021fairdrop} alters graph topology to reduce homophily related to sensitive attributes.
EDITS \citep{dong2022edits} goes further by adjusting both graph topology and node attributes based on the distance among demographic groups.
In-processing methods represent another pipeline that revises the model training process to achieve more fair outcomes.
For instance, 
FairGNN \citep{dai2021say} integrates an adversary to achieve fair outputs for node classification with limited sensitive attributes, while Graphair \citep{ling2022learning} seeks to learn fair representations by automated graph data augmentations.
FairVGNN \citep{wang2022improving} tackles the sensitive attribute leakage caused by feature propagation in GNNs by automatically learning from fair views.
Additionally, FairGKD \citep{zhu2023devil} investigates the fairness performance in different training strategies and uses distilled knowledge from partial data training to enhance model fairness.
However, directly applying the above methods for GAD poses significant challenges. 
A primary issue is the overlap between anomalous characteristics and sensitive attributes, which complicates the debiasing process.
Besides, these methods are typically designed for or validated on supervised tasks, while the lack of ground truth is a fundamental issue in anomaly detection.

\subsection{Fair Representation Learning}
Fair representation learning has shown great success in learning representations free from sensitive information while maintaining downstream task-related information for decision-making~\citep{zemel2013learning,liu2022fair}.
Disentangled representation learning provides a novel perspective in fair representation learning, enabling the simultaneous maintenance of sensitive-relevant and -irrelevant information, which are separated into independent subspaces. 
While disentangled fair representation learning has shown promise in benefiting fairness in image classification \citep{creager2019flexibly,kim2021counterfactual,oh2022learning}, its applicability on unsupervised GAD is non-trivial.
The first challenge involves effectively encoding node attributes and graph topology into disentangled i.i.d. representations while preserving the information essential for anomaly detection.
An additional challenge lies in achieving fair GAD based on sensitive-irrelevant representations.
In prior studies, disentangled sensitive-irrelevant representations are directly used for supervised downstream tasks \citep{creager2019flexibly,oh2022learning}.
However, in unsupervised GAD, where labels are not available, the reconstruction error serves as an effective criterion for decision-making.
Considering the correlation between sensitive attributes and other attributes, as well as the connections influenced by sensitive attributes, direct reconstruction of the original graph may result in biased decisions.

\section{Conclusion}
In this paper, we propose \method, a novel disentangle-based framework for fair graph anomaly detection, aiming to balance fairness and utility in decision-making.
To the best of our knowledge, \method is the first method in enhancing fairness in the task of graph anomaly detection. 
\method first introduces disentangled representation learning to capture informativeness yet sensitive-irrelevant representations, thereby mitigating societal bias associated with sensitive attributes within the input graph. 
Furthermore, to alleviate discriminatory decisions in anomaly detection, \method reconstructs input attributes from the sensitive-irrelevant representations and implements a constraint on the correlation between reconstruction error and predicted sensitive attributes. 
The effectiveness of \method has been substantiated through extensive experiments on real-world datasets, demonstrating its superiority over several baselines regarding accuracy and fairness. We discuss limitations and future work in \autoref{app:limitations}.

\subsubsection*{Acknowledgments}
This research is supported by a grant from the National Key Research and Development Program of China (No.2022YFF0711801) and the CAS 145 Informatization Project CAS-WX2022GC-0301. This work is supported in part by NSF under grants III-2106758 and POSE-2346158. Jianjun Yu is the corresponding author.

\bibliography{main}
\bibliographystyle{tmlr}

\appendix

\section{Pseudo Code} \label{sec:training}
The overall training process of \method is presented in Algorithm~\ref{alg}.
Firstly, given an attributed graph $\mathcal{G}$, we aim to train a variational graph autoencoder, in which the disentangled encoder $f_{e}$ can separate sensitive-irrelevant representations while maintaining information for reconstructing $\mathcal{G}$ by the decoder $f_{a}$ and $f_{x}$ (Line 4-6). 
We encourage the disentanglement with a learnable adversary $g_{\omega}$ optimized by $\mathcal{L}_{adv}$ (Line 8-9). 
The optimization of $f_{e}$, $f_{a}$ and $f_{x}$ and that of $g_{\omega}$ are conducted adversarially.
Next, the frozen $f_{e}$ captures sensitive-irrelevant node representations (Line 14), and the decoder $f_{\phi}$ solely reconstructs attributes (Line 15). 
We utilize reconstruction error as the anomaly score and constrain the correlation between reconstruction error and predicted sensitive attributes (Lines 16-19).
Finally, only $f_{\phi}$ will be updated by $\mathcal{L}_{ad}$ in the anomaly detection phase.
\let\AND\undefined
\begin{algorithm}[h] 
    \renewcommand{\algorithmicrequire}{\textbf{Input:}}
    \renewcommand{\algorithmicensure}{\textbf{Output:}}
    \caption{\method algorithm}
    \label{alg}
    \begin{algorithmic}[1]
        \REQUIRE Graph $\mathcal{G}=(\mathcal{V}, \mathcal{E})$, adjacency matrix $\mathbf{A}$, attribute matrix $\mathbf{X}$, sensitive attribute matrix $\mathbf{S}$, GNN encoder $f_{e}$, attribute decoder $f_{x}$, structure decoder $f_{a}$, binary adversary $g_{\omega}$, linear attribute decoder $f_{\phi}$, training iteration $T$;
        \ENSURE Anomaly scores $\mathbf{o}$ 
        \STATE \textit{// Disentangled representation learning phase;}
        \STATE Initialize $f_{e}, f_{a}, f_{x}, g_{\omega}$;
        \WHILE{not converged}
            \STATE $\mathbf{Z}_x, \mathbf{Z}_s \leftarrow  f_{e}(\mathbf{X},\mathbf{A})$;
            \STATE $\hat{\mathbf{A}}, \hat{\mathbf{X}} \leftarrow f_{a}(\mathbf{Z}_x,\mathbf{Z}_s), f_{x}(\mathbf{Z}_x,\mathbf{Z}_s)$;
            \STATE Calculate $\mathcal{L}_{total}$ according to \autoref{eq:total_loss};
            \STATE Update $f_e, f_a, f_x$ by gradient descent using $\mathcal{L}_{total}$;
            \STATE Calculate $\mathcal{L}_{adv}$ according to \autoref{eq:adv_loss};
            \STATE Update $g_\omega$ by gradient descent using $\mathcal{L}_{adv}$;
        \ENDWHILE
        \STATE \textit{// Anomaly detection phase;}
        \STATE Freeze $f_{e}$ and initialize $f_{\phi}$;
        \FOR{$t=1,2,\dots,T$}
            \STATE $\bar{\mathbf{Z}}_x, \mathbf{Z}_s \leftarrow  f_{e}(\mathbf{X},\mathbf{A})$;
            \STATE $\tilde{\mathbf{X}} \gets f_{\phi}(\bar{\mathbf{Z}}_x, \tilde{\mathbf{Z}}_s)$;
            \STATE Calculate anomaly score $\mathbf{o}$ according to \autoref{eq:a_score};
            \STATE Calculate $\mathcal{L}_{ad}$ according to \autoref{eq:ad_loss};
            \STATE Update $f_{\phi}$ by gradient descent using $\mathcal{L}_{ad}$;
        \ENDFOR
        \RETURN $\mathbf{o}$;
    \end{algorithmic}
\end{algorithm}

\section{Complexity Analysis}
\label{sec:complexity}

We analyze the time and space computational complexity of \method. Let $N$, $E$, $d$ and $h$ denote the number of nodes, edges, attribute dimensions and hidden dimensions, respectively. \\
In the disentangled representation learning phase, the encoder $f_e$ exhibits a complexity of $O(Ed)$, while the decoders $f_a$ and $f_x$ exhibit a complexity of $O(N^2)$ for adjacency matrix reconstruction. The adversary $g_\omega$ operates at $O(Nh)$. For loss computations, the variational reconstruction loss $\mathcal{L}_{vae}$ exhibits $O(N^2)$ complexity for adjacency matrix reconstruction and $O(Nd)$ for attribute reconstruction. The disentanglement loss $\mathcal{L}_{dis}$, predictive loss $\mathcal{L}_{pre}$, and adversarial loss $\mathcal{L}_{adv}$ each exhibit $O(Nh)$ complexity. 
In the anomaly detection phase, the frozen encoder $f_e$ maintains $O(Ed)$ complexity, while the MLP-based decoder $f_\phi$ exhibits $O(Nh)$ complexity. Both the correlation constraint loss $\mathcal{L}_{corr}$ and reconstruction loss $\mathcal{L}_{rec}^X$ exhibit $O(Nh)$ complexity. Thus, the total time complexity is $O(N^2 + Ed)$. The space complexity of \method is dominated by the storage of the dense adjacency matrix $O(N^2)$, node feature matrix $O(Nd)$, and model parameters with intermediate results $O(Nh)$, resulting in a total space complexity of $O(N^2 + Nd + Nh)$. \\
To enhance scalability, we implement a sparse version that computes reconstruction only for existing edges and approximates the global statistics using row/column means, instead of reconstructing the full adjacency matrix. We also replace dense matrix operations with sparse operations in loss computations. 
This optimization reduces the complexity of the decoder $f_a$ to $O(Eh)$ for structure reconstruction. Besides, the complexity of $\mathcal{L}_{vae}$ decreases to $O(E)$ for structure and maintains $O(Nd)$ for attribute reconstruction. Thus, the overall time complexity is reduced to $O(Ed + Nd)$.
The space complexity decreases significantly to $O(E + Nd + Nh)$ by utilizing sparse matrix representations and eliminating dense adjacency matrices.

\section{Datasets} \label{app:datasets}
We employ three real-world datasets for fair GAD, which provide both real sensitive attributes and ground-truth labels for GAD. 
In Reddit and Twitter datasets, the sensitive attribute is the political leaning of users, while the anomaly label is assigned to misinformation spreaders. 
The Credit dataset focuses on payment default detection, with age as the sensitive attribute.
Details of these datasets are summarized in \autoref{tab:dataset}. 

\begin{itemize}[leftmargin=*]
    \item \textbf{Reddit}~\citep{neo2024towards} contains 110 politically oriented subreddits, encompassing all historical postings within these forums. It also includes all historical posts from several active discussion participants. A relational graph was constructed by linking users who posted in the same subreddit within 24 hours.
    \item \textbf{Twitter}~\citep{neo2024towards}  is conducted on 47,712 users with historical posts, user profiles, and follower relationships. The user information, such as the organization status, was inferred using the M3 System from user profiles and tweets. The key account metrics and averaged post embeddings were combined to form node features. The network structure was established based on follower relationships between users.
    \item \textbf{Credit}~\citep{agarwal2021towards}  contains 30,000 individuals with features like education, credit history, age, and features derived from their spending and payment patterns. Two nodes are connected if their similarity exceeds 70\% of the maximum similarity between all node pairs, measured using Minkowski distance.
\end{itemize}

\section{Baselines} \label{app:baselines}
In this subsection, we introduce the baselines employed in our experiments, including GAD methods (i.e., DOMINANT, CoLA, CONAD, and VGOD), Fairness Regularizers (i.e., FairOD, Correlation, and HIN), and Graph Debiasers(i.e., FairWalk and EDITS).
\begin{itemize}[leftmargin=*]
    \item \textbf{DOMINANT}~\citep{ding2019deep} devises a GCN-based autoencoder to detect anomalies by reconstructing node attributes and graph structure. The attribute decoder is the reverse structure of the encoder and the structure decoder is applied by dot product. Anomalous nodes are evaluated by reconstruction errors.
    \item \textbf{CoLA}~\citep{liu2021anomaly} is a contrastive self-supervised learning framework for anomaly detection. It conducts a node-subgraph contrast to capture anomalies that are dissimilar from their local neighbors. Anomalous nodes are evaluated by the agreement between each node and its neighboring subgraph with a GNN-based model.
    \item \textbf{CONAD}~\citep{xu2022contrastive} introduces a contrastive learning framework that leverages human knowledge through data augmentation to enhance anomaly detection capabilities. 
    It employs a Siamese graph neural network with a contrastive loss to encode both the modeled knowledge and the original attributed networks. Anomalous nodes are evaluated by reconstruction errors.
    \item \textbf{VGOD}~\citep{huang2023unsupervised} proposes a variance-based framework that combines a variance-based model for structural outlier detection with an attribute reconstruction model for contextual outlier detection. It achieves balanced detection performance between structural and contextual outliers while addressing data leakage issues present in existing injection-based approaches.
    \item \textbf{FairOD}~\citep{shekhar2021fairod} is a fairness-aware outlier detector on independent and identically distributed (i.i.d.) data. It devises a regularization term to prompt the fairness of demographic parity by minimizing the reconstruction errors $\mathbf{o}$ and the sensitive attributes $\mathbf{S}$. 
    \begin{equation}
        \mathcal{L}_{DP}^{FairOD} = \left | \frac{\left ( \sum_{i=1}^n {o}_i - \mu_{o} \right ) \left ( \sum_{i=1}^n {s}_i - \mu_{s} \right )}{\sigma_{o} \sigma_{S}} \right |
    \end{equation}
    where $\mu_{o}$ and $\mu_{s}$ represent the means, while $\sigma_{o}$ and $\sigma_{s}$ denote the corresponding standard deviations of $\mathbf{o}$ and $\mathbf{S}$, respectively. 
    Besides, it utilizes an approximation of Discounted Cumulative Gain (DCG) to enforce the group fidelity.
    \begin{equation}
        \mathcal{L}_{ADCG}^{FairOD} = \sum_{s\in \{0,1 \}} \left ( 1-\sum_{\{v_i:s_i=s \}} \frac{2^{{o}_i^{base} }-1}{\mathrm{DNM}_s} \right ), 
    \end{equation}
    where \( \mathrm{DNM}_s=\log_2\left ( 1+ \sum_{\{v_j:s_j=s \}} \sigma({o}_j - {o}_i) \right ) \cdot IDCG_{s} \) and \( {IDCG}_{s} = \sum_{j}^{|\{v_j:s_j=s \}|} \left( {2^{{o}_j^{base} }-1}/{\log_2(1+j)} \right) \). Here, ${o}_i^{base}$ is the reconstruction error of $v_i$ in the base model and $\sigma(\cdot)$ is the Sigmoid function.
    The overall loss of the model equipped with FairOD regularizer can be calculated by $\mathcal{L} = \mathcal{L}_{base} + \lambda \mathcal{L}_{DP}^{FairOD} + \gamma  \mathcal{L}_{ADCG}^{FairOD}$,
    where $\lambda$ and $\gamma$ are weight parameters.
    \item \textbf{Correlation}~\citep{shekhar2021fairod} is an implementation of FairOD, which measures the correlation between sensitive attributes $\mathbf{S}$ and reconstruction errors $\mathbf{o}$ using the cosine similarity.
    \begin{equation}
        \mathcal{L}^{Corr} = \left| \frac{\mathbf{o} \cdot \mathbf{S}}{\sqrt{(\mathbf{o} \cdot \mathbf{o}) (\mathbf{S} \cdot \mathbf{S})}}  \right|,
    \end{equation}
    where $(\cdot)$ represents the dot product of two vectors.
    The overall loss of the model equipped with the Correlation regularizer can be calculated by $\mathcal{L} = \mathcal{L}_{base} + \lambda \mathcal{L}^{Corr}$.
    \item \textbf{HIN}~\citep{zeng2021fair} focuses on fair representation learning for heterogeneous information networks. The demographic parity-based fairness-aware loss function is calculated by:
    \begin{equation}
        \mathcal{L}_{DP}^{HIN} = \sum_{k\in\{0,1\}} \Bigg(  \frac{\sum_{\{v_i:s_i=1\}}P(\hat{y}_i=k)}{|{\{v_i:s_i=1\}}|} 
         - \frac{\sum_{\{v_i:s_i=0\}}P(\hat{y}_i=k)}{|{\{v_i:s_i=0\}}|} \Bigg)^2,
    \end{equation}
    where $P(\hat{y}_i=1)$ denotes the predicted probability of $v_i$ to be identified as an anomalous node.
    The equal opportunity-based fairness-aware  loss is calculated by:
    \begin{equation}
        \mathcal{L}_{EO}^{HIN} = \sum_{k\in\{0,1\}} \Bigg(\frac{\sum_{\{v_i:s_i=1,y=k\}}Pr(\hat{y}_i=k)}{|{\{v_i:s_i=1,y=k\}}|} 
        - \frac{\sum_{\{v_i:s_i=0,y=k\}}Pr(\hat{y}_i=k)}{|{\{v_i:s_i=0,y=k\}}|} \Bigg)^2.
    \end{equation}
    As the calculation of $\mathcal{L}_{EO}^{HIN}$ requires task-related labels and anomaly detection tasks typically lack ground truth labels, we use $\mathcal{L}_{ADCG}^{FairOD}$ as a replacement to enhance the fairness in equal opportunity.
    Thus, the modified overall loss of the model equipped with HIN regularizer can be calculated by:
    \begin{equation}
        \mathcal{L} = \mathcal{L}_{base} + \lambda \mathcal{L}_{DP}^{HIN} + \gamma  \mathcal{L}_{ADCG}^{FairOD}.
    \end{equation}
    \item \textbf{FairWalk}~\citep{rahman2019fairwalk} introduces a fairness-aware embedding method that generates node embeddings by considering sensitive attributes and the topology of the graph.  It enhances the general random walk algorithm to capture more diverse neighborhoods, thereby producing embeddings that exhibit reduced bias.
    \item \textbf{EDITS}~\citep{dong2022edits} is a graph debiasing method for attributed graphs, mitigating bias present in both graph topology and node features. It minimizes the approximated Wasserstein distance between the distributions of different groups for any attribute dimension to enhance group fairness.
\end{itemize}

\section{Implementation Details} \label{app:implementation}
For DOMINANT, CoLA, CONAD and VGOD, we use the code and default hyper-parameters provided by PyGOD\footnote{\url{https://pygod.org}}~\citep{liu2022pygod}.
For FairWalk\footnote{\url{https://github.com/urielsinger/fairwalk}} and EDITS\footnote{\url{https://github.com/yushundong/EDITS}}, we implement them using the code published by their authors.
For FairOD, HIN, and Correlation, we implement the code provided by FairGAD\footnote{\url{https://github.com/nigelnnk/FairGAD}}.

\section{Limitations and Future Work} \label{app:limitations}
First, DEFEND encounters difficulties when protected attributes are highly correlated with other features, potentially resulting in information loss during disentanglement. Additionally, DEFEND relies on adversarial learning to approximate the Total Correlation penalty used for disentanglement. This dependency introduces potential convergence instability issues that might compromise the robustness of learned representations \citep{oh2022learning}. We plan to extend the framework to handle multiple and continuous sensitive attributes simultaneously, which would enhance its practical applicability. 

\end{document}